\documentclass[11pt, a4paper, logo, onecolumn]{googledeepmind}

\usepackage[titles]{tocloft}
\usepackage{setspace}
\usepackage[authoryear, sort&compress, round]{natbib}
\title{Multi-Actor Generative Artificial Intelligence as a Game Engine}

\correspondingauthor{}

\reportnumber{} 

\renewcommand{\today}{July 2025}

\author[1]{Alexander Sasha Vezhnevets}
\author[1]{Jayd Matyas}
\author[1]{Logan Cross}
\author[1]{Davide Paglieri}
\author[1]{Minsuk Chang}
\author[1, 2]{William A. Cunningham}
\author[1]{Simon Osindero}
\author[1]{William S.~Isaac}
\author[1]{Joel Z. Leibo}

\affil[1]{Google DeepMind}
\affil[2]{University of Toronto}

\begin{document}

\begin{abstract}
    Generative AI can be used in multi-actor environments with purposes ranging from social science modeling to interactive narrative and AI evaluation. Supporting this diversity of use cases---which we classify as Simulationist, Dramatist, and Evaluationist---demands a flexible scenario definition framework. We argue here that a good approach is to take inspiration from tabletop role-playing games (TTRPGs), where a Game Master (GM) is responsible for the environment and generates all parts of the story not directly determined by the voluntary actions of player characters. We argue that the Entity-Component architectural pattern is useful here. In such a system, the GM is not a hardcoded computer game but is itself a configurable entity, composed of components just like any other actor. By design, the approach allows for a separation between the underlying implementation details handled by an engineer, the creation of reusable components, and their  composition and configuration managed by a designer who constructs entities from the components. This separation of concerns is instrumental for achieving rapid iteration, maintaining modularity, and ultimately to ensure scalability. We describe the ongoing evolution of the Concordia library in terms of this philosophy, demonstrating how it allows users to effectively configure scenarios that align with their specific goals.
\end{abstract}

\maketitle

{
\singlespacing
\tableofcontents
}

\section{Introduction}

The space of fruitful application for ``agentic'' generative artificial intelligence (AI) is rapidly expanding beyond single-actor systems into complex multi-actor environments \citep{park2023generative, li2023camel, vezhnevets2023generative, guo2024large, han2024llm, sunehag2025simulation, yang2025survey}. These systems are being developed for a wide range of purposes: generating compelling narratives \citep{wu2024role, mirowski2023co}, modeling tools for social and psychological science \citep{anthis2025llm, kozlowski2024silico, park2024generative, horton2023large, grossmann2023ai, hewitt2024predicting}, sandboxes for problem solving teamwork  \citep{shao2024collaborative, gong2023mindagent, zhang2023building, zeng2024autodefense}, evaluation frameworks \citep{zhou2023sotopia,guertler2025textarena} and synthetic data sources \citep{duenez2023social, silver2025welcome, motwani2024malt, liu2025spiral}. This diversity in application domains suggests it will be useful to carefully consider underlying architecture and design philosophies to make sure flexibility over all the relevant use cases can be maintained without adding too much complexity.

To better frame the diverse applications of these systems, we propose a typology of user motivations taking inspiration from game design theory (specifically \cite{edwards2004system}). We delineate three distinct motivations for defining generative multi-actor scenarios: (A) Evaluationist, (B) Dramatist, and (C) Simulationist. The Evaluationist seeks to create controlled environments to benchmark and compare AI capabilities, their most important requirement is for the system to be a fair and rigorous testbed (Section~\ref{section:evaluationist}). The Dramatist aims to generate compelling, coherent narratives and rich character interactions, using the system as an interactive storytelling engine (Section~\ref{section:dramatist}). Finally, the Simulationist uses the system as a ``pocket universe'' to model and understand real-world social or causal dynamics with as much fidelity as possible (Section~\ref{section:simulationist}). We also discuss synthetic data generation for training further AI systems as a cross-cutting concept with connection to all three motivations (Section~\ref{section:syntheticData}). Such datasets can be used as curricula to train new AI systems. In this view, a Dramatist configuration becomes a `story factory', capable of generating endless narrative examples to forge more creative and empathetic models. Data from Simulationist scenarios can be used to imbue an AI with greater understanding of complex social and physical dynamics. Likewise, an Evaluationist framework can serve as a diagnostic engine, producing fine-grained behavioral data across controlled conditions to enable precise, quantitative comparisons between different models. Which motivation becomes most salient depends on the ultimate goal of the work (Evaluationist, Dramatist, or Simulationist). We will argue that the requirements of effectively serving any of these distinct goals can be achieved in single library of sufficient flexibility.

The original Concordia \citep{vezhnevets2023generative} was conceived as a generative agent-based modeling library (GABM) composed of two  generative parts: a model of individual actors and a model of the environment, which was termed the Game Master (GM)---both based on large language models. The GM, inspired by the storyteller role in tabletop role-playing games \citep{gyraxcook1989dnd}, was responsible for generating the setting and context for social interactions of the individual actors. The ``players'' in the TTRPG context are identified with the individual decision makers of agent-based modeling.

The subsequent development of Concordia maintains the inspiration from table-top role-playing games while evolving toward a technical design philosophy aligned with modern game engines like Unity 3D \citep{unity3d}. The core of this evolution is the principled adoption of the Entity-Component architectural pattern. An Entity is a simple, unique identifier—essentially a container. A Component is a reusable module of data that defines a specific aspect of an entity, such as its memory, physical state, or goals. An entity's properties and behaviors emerge from the specific collection of components attached to it. While entities often represent individual agents, the flexibility of this pattern means an entity can also represent a collective, such as a company, a government agency, a social group, or a legal body like a court. Their Components would then pertain to the collective attributes and functions of that organization, like accounting ledgers and court proceedings.

In terms of Concordia's TTRPG interpretation, both the players and the game master are entities. An entity's functionality is defined by its components.

The Entity-Component architectural choice is the technical foundation for our design philosophy, which separates the concerns of the `engineer' and the `designer'. The engineer uses the Concordia framework to build stable, reusable, and testable components (e.g., a new planning algorithm or a specific memory model\footnote{Components often use RAG internally, see \cite{vezhnevets2023generative}. This is similar to \cite{park2023generative}.}). The designer uses existing entities and components as building blocks to craft scenarios, often without writing new component-level code. 

We argue that this separation is very helpful in a multi-actor generative AI platform. It allows a user to focus entirely on their design goal---be it Evaluationist, Dramatist, or Simulationist---by configuring existing components, while another developer can extend the platform's fundamental capabilities. Even if there is only a single person playing both roles, the separation of concerns between design and engineering is still helpful, both for enabling rapid experimentation at the design level and stable, systematic development at the engineering level.

\section{Entities, Components, Engines, and Game Design}

The Entity-Component architectural pattern, a cornerstone of modern game development \cite{nystrom2014game,gregory2018game,raffaillac2019polyphony}, provides a powerful and flexible foundation for building multi-actor generative AI systems that can be effectively tailored to different design goals. An Entity-Component framework utilizes composition instead of inheritance, as entities are not defined by a rigid class structure. Instead, they are independent lightweight objects with unique identifiers. Their behaviors and properties are determined by the components attached to them (i.e. an entity is a box of components with a name). The engine (i.e. a ``system'' in strict entity-component-system terminology) is responsible for processing entities by calling functions such as `observe' and `act' on all entities with components implementing those functions. 

Components are created by combining Python code with LLM calls. This allows for maximal flexibility and expressivity. When the designer knows how to code a specific feature they are welcome to do so; simultaneously, other features in the same environment can be implemented by asking the GM storyteller LLM to implement them. Both kinds of functionality generally exist in the same environment. You can give the GM freedom to create whatever its LLM comes up with, or you can constrain it to stay very close to established, and hard-coded, guardrails---and everything in between.

There are two main calls on an entity: \texttt{observe} and \texttt{act}.

When \texttt{observe} is called, this triggers \texttt{preobserve} and \texttt{postobserve} functions to be called on every component to process observations for each entity. When \texttt{act} is called, each component plays one of two distinct roles:
\begin{itemize}
    \item \textbf{Context}---These can be numerous and diverse. They initially process the action request in parallel, each contributing information or updating its internal state. Their outputs are aggregated to form a comprehensive contextual understanding for the entity. After an action is determined, these components are again engaged to react to the outcome. For example, a SelfReflection context component asks ``What kind of person am I?'' and feeds the LLM's response to subsequent components (as described in \cite{vezhnevets2023generative} and \cite{leibo2024theory}). 
    \item \textbf{Acting}---Each entity has exactly one component designated as the Acting component. This restriction to a single acting component reflects a basic assumption that an actor takes exactly one action per step. Note that the amount of time passing between steps is entirely arbitrary so this is not a restrictive assumption. Knowing there will be at most one action per entity per step makes it natural to view the Acting component as an aggregator of information received from a set of Context components. It receives information from calling \texttt{preact} on all Context Components, and an action request from the engine (i.e. ultimately from another entity, such as the GM). The Acting component's unique responsibility is to synthesize all this input to decide and return the entity's final, singular action (for the current timestep). This clear division of labor---many components providing and reacting to context, and one component deciding the action---underpins the system's flexibility. It allows testing of variable decision-making strategies by interchanging the Acting component while retaining the same suite of Context Components. After the Acting component returns the action, \texttt{postact} is called on all components.
\end{itemize}

In practice, when implementing a Concordia component, what you actually do is you implement \texttt{preobserve}, \texttt{postobserve}, \texttt{preact}, and \texttt{postact} (or any subset of these). It is very common to implement either the observation methods or the action methods and rare to implement both kinds on the same component.

The modularity of components allows for the easy creation of diverse entities with different functionality by mixing and matching components. This contrasts with traditional object-oriented approaches where creating new types of actors with slightly different behaviors can lead to complex and brittle inheritance chains. For generative AI agents, this is particularly beneficial. An agent's ``mind'' can be composed of various components: a Memory component for storing past experiences, a Planning component that uses a large language model to generate goals, and a Beliefs component to represent its understanding of the world. Similarly, an organizational entity could be composed of components representing its departments, policies, and internal communication structures. Different agents can be endowed with different cognitive architectures by simply equipping them with different sets of components. This modularity makes the framework very scalable. It can support scenarios ranging from a simple dialogue between two actors in a chat room to a complex economic simulation with thousands of agents and external events like natural disasters.

The flexibility of this architectural pattern also extends to the Game Master (GM) within the Concordia framework. The GM is also an entity and can be customized with components, just like player entities (actors). This allows the GM's role and logic to be tailored to the specific needs of the multi-actor system, whether that be enforcing strict evaluation protocols, guiding narrative development, or maintaining causal consistency. The use of prefabs, or pre-configured collections of components (which can be cloned and subsequently modified), can further streamline the process of creating and customizing both actors and GMs. 

The Concordia framework also accommodates diverse interaction dynamics through several game engines. A simultaneous engine allows all actors to act in parallel within the same time step, useful for modeling scenarios like market trading where all actors submit their buy/sell orders at the same time. In contrast, a sequential engine enforces turn-based interactions---such as dialogue---with an order that is either predetermined or dynamically generated by the Game Master. The framework also supports fully asynchronous operation, which is ideal for modeling environments like social media where users can act at any time, independent of others. This flexibility can also be used to speed up or scale up simulations via concurrency.

One goal in designing these language features, both here and in the game engines like \cite{unity3d} that inspired us, was to allow for a differentiation in role between game designers and game engineers. Of course, actual differentiation of roles is optional (and unlikely in small teams). Nevertheless, the modularity enforced by designing around the designer/engineer dichotomy is helpful for everyone in terms of facility for defining robust and flexible scenarios. The distinction between engineer and designer in Concordia reflects a deliberate prioritization: the system is built to make design work possible without constant engineering support. Users should be able to set up complex meaningful scenarios without needing to write new component-level code. This mirrors workflows in modern game engines like Unity, which are structured to let designers build and test systems by combining multiple parts \citep{fullerton2008game} In the context of generative multi-actor systems, this makes Concordia especially well-suited for rapid hypothesis testing, scenario exploration, and iterative modeling of social dynamics.

\section{Mapping the space of potential game/simulation design objectives}
\label{section:threeMotivations}

Taking inspiration from ``System Does Matter\footnote{\url{http://www.indie-rpgs.com/_articles/system_does_matter.html}}'' \citep{edwards2004system}, we discuss the varieties of applications and kinds of multi-actor\footnote{Note that we have started using the term `actor'' as opposed to the term ``agent'' since we think it is likely that the term ``agent'' will soon be synonymous with a category of tech product and rapidly lose its original meaning, the same semantic transformation that recently occurred for the word ``hallucination''.} generative AI approaches. Just as there are three kinds of tabletop role-playing gamers (or three motivations for playing tabletop role-playing games (TTRPG), there are analogous categories for the users of multi-actor generative AI system design libraries. According to \cite{edwards2004system}, the categories for TTRPGs are: (1) Gamist, (2) Narrativist, and (3) Simulationist. The Gamist wants the system to be fair and balanced, and to afford the possibility of victory while maintaining danger of defeat. The Narrativist wants the result to be a good story. The Simulationist wants a ``little pocket universe without fudging'' \cite{edwards2004system}---they seek realistic projection from initial conditions.

We will argue here that users of multi-actor generative AI also have a range of different motivations and that a specific classification schema is helpful for understanding their diversity. It is important, however, to clarify how we are adapting this analogy. Whereas Edwards used his model to categorize system types and their internal mechanics, we use this framework to describe the different goals a user might bring when configuring an environment. The framework characterizes user intent, not a prescription for how a system ought to be designed. This distinction is key to our argument that a flexible platform can be configured to serve any of these user goals.

With that in mind, we find it helpful to consider the motivations for using multi-actor generative AI as belonging to one of the following types: (1) Evaluationist (analogous to \cite{edwards2004system}'s Gamist), (2) Dramatist (analogous to \cite{edwards2004system}'s Narrativist), and (3) Simulationist (analogous to \cite{edwards2004system}'s category of the same name). The Evaluationist wants to evaluate the performance of AI systems, determine which are better or worse and in which applications and contexts, the Dramatist wants the multi-actor system to function well as a story generator, and the Simulationist wants the multi-actor system to function as a modeling tool for forecasting and social science research. Our \cite{edwards2004system}-inspired framework is also related to another influential game design framework called MDA \citep{hunicke2004mda}, which separates game systems into Mechanics (implementation-level rules), Dynamics (emergent behavior during execution), and Aesthetics (intended outcomes or system-level goals).

Generative AI also has a fourth motivation which may perhaps be its own primary category, or may instead be considered a background objective that permeates all three primary objectives and colors them: this is the objective of creating synthetic training data (see Section~\ref{section:syntheticData}). We think it is better to view synthetic data generation as a background goal because the generated  data will itself always have to be data suitable for a specific purpose. If the purpose is to improve the AI system's fiction writing and understanding then the data-generating multi-actor system will be constructed in accord with the Dramatist motivation. If the purpose is to improve the AI system's realistic understanding of social or physical causality then the motivation for the data generating multi-actor system is Simulationist.

\subsection{The Evaluationist Perspective}
\label{section:evaluationist}

The Evaluationist approach to multi-actor generative AI systems parallels \cite{edwards2004system}'s Gamist motivation in TTRPGs. Where the Gamist seeks fair contests with opportunities for strategic victory, the Evaluationist approaches multi-actor systems as frameworks for assessment and comparison. We have taken this perspective in benchmark-oriented research such as the Concordia Contest \citep{smith2025evaluating}, which evaluated the generalization capabilities of LLM-based actors in mixed-motive scenarios.

For Evaluationist users, the primary objective is clear: to determine which AI systems perform better across specified dimensions and contexts. This requires environments that provide a ``fair playing field'' with well-defined metrics for success. Systems designed with this motivation prioritize balanced starting conditions, consistent evaluation protocols, and reproducible outcomes. The focus is not on narrative coherence or simulation fidelity per se, but rather on creating conditions that allow for meaningful comparison between different agent implementations.

Evaluationist systems typically feature:
\begin{itemize}
\item \textbf{Standardized scenarios}---Carefully calibrated environments that present consistent challenges across multiple evaluation runs
\item \textbf{Clear success metrics}---Quantifiable measures of performance that allow for unambiguous ranking of different approaches
\item \textbf{Controlled variability}---Strategic introduction of novel elements to assess generalization capabilities
\item \textbf{Cross-play mechanics}---Methods for evaluating how agents perform when interacting with different partner populations
\end{itemize}

In the Concordia NeurIPS competition \citep{smith2025evaluating}, the evaluation emphasized zero-shot generalization---the ability of agents to transfer strategies to novel social contexts. This methodological approach mirrors principles from supervised learning research, where performance on previously unseen data is considered the gold standard for evaluation. The contest structure, with its distinct development and evaluation phases, is an example of the way in which adopting the Evaluationist motivation has implications for system design.

Furthermore, the Evaluationist can use this framework not just to test overall model capability, but to diagnose why an agent succeeds or fails. Because agents are composed of modular components that form a structured chain of thought, a researcher can build, test, and refine specific cognitive architectures (`decision logics' in the terminology of \cite{leibo2024theory}). For example, if an agent consistently fails at a negotiation task, the Evaluationist can swap out its Planning component or add a new SocialReasoning component to investigate and fix the specific point of failure in its decision-making.

\subsection{The Dramatist Perspective}
\label{section:dramatist}

In contrast to the Evaluationist, the Dramatist views multi-actor generative AI systems primarily as narrative engines. This motivation aligns with \cite{edwards2004system}'s Narrativist player, who is ``satisfied if a roleplaying session results in a good story.'' For a user with a Dramatist goal, the central concern is not benchmarking performance but generating compelling narratives through the interactions of multiple AI actors. Multi-actor systems for narrative-driven computer games are often of this type.

From the designer's perspective, a system built for a Dramatist purpose will prioritize narrative coherence, emotional resonance, and dynamic character development over standardized evaluation. This leads to a focus on features such as:
\begin{itemize}
\item \textbf{Rich character models} -- Detailed personas with well-defined goals, values, and relationships, often constructed by composing multiple components
\item \textbf{Narrative-driven environments} -- Settings designed to elicit dramatically interesting interactions
\item \textbf{Flexible resolution mechanisms} -- Systems that prioritize narrative satisfaction over procedural consistency
\item \textbf{Emergent story arcs} -- Frameworks that allow for the development of compelling narrative trajectories without predetermined outcomes
\end{itemize}

Achieving these narrative goals is a quintessential task for the game designer. Within the Concordia framework, the designer can leverage the modularity of the Entity-Component architecture to orchestrate the story. For example, the Game Master (GM) entity can be configured with specific components to function as a "narrative director." A designer might give the GM components that create distinct protagonist and antagonist roles among the actors, and then manage information asymmetry between them. An antagonist character, for example, could be given knowledge of future plot points, such as events in a third act, while the protagonist remains unaware, thereby heightening dramatic tension. Alternatively, the GM can be configured to facilitate a dynamic, player-driven experience, essentially creating an open-ended ``choose-your-own-adventure'' storyline. \cite{jenkins2004game} frames this as a distinction between \textit{embedded} and \textit{emergent} narrative structure---the GM can either lay out key events in advance or let narrative arcs take shape through interaction. Concordia supports both approaches, depending on how the scenario is configured. This model is similar to that of games like AI Dungeon\footnote{\url{https://aidungeon.com/}}, where the player's freeform text choices direct an LLM to generate worlds, characters, and scenarios in real-time. This approach is part of a broader effort to use generative AI for creating dynamic storylines that adapt to player choices, moving beyond traditional branching narratives toward truly emergent storytelling \citep{korsah2025impact, gallotta2024large}.

\subsection{The Simulationist Perspective}
\label{section:simulationist}

The Simulationist approach parallels its TTRPG counterpart, seeking to create ``a little pocket universe without fudging'' \citep{edwards2004system}. For these users, the goal is to model real-world processes with enough fidelity to generate insights into complex social, economic, or physical phenomena \citep{anthis2025llm}. This perspective directly addresses a long-standing challenge in the social sciences. For decades, Agent-Based Models (ABMs) have been a vital tool, but their power has been limited by the need for simplifying assumptions about agents and their environments \citep{poteete2010working}. Often, agents are modeled as simple, rational reward-maximizers, and environments are reduced to a few variables, which limits the kinds of phenomena that can be studied \citep{schill2019more}.

The goal of a Simulationist using Concordia is to try to realize the original promise of ABMs. This is achieved in two primary ways. First, by populating simulations with more psychologically and sociologically plausible actors. Unlike agents in traditional ABMs or even many multi-agent reinforcement learning systems (see \cite{hertz2025beyond}), Concordia models can be defined without any need for a scalar reward signal. Entities in the simulations act via ``pattern completion'' based on their identity and experience \citep{leibo2024theory}. This allows for the modeling of nuanced social dynamics---such as the spread of identity-based gossip or how arbitrary social norms are negotiated---that were previously too difficult to capture computationally. Second, Concordia allows for greater environmental fidelity, as the GM can generatively manage open-ended world details, freeing the designer to code only the specific environmental mechanics critical to their research question.

A designer with a Simulationist motivation is not adhering to a rigid structure, but instead focuses on a set of priorities tailored to their research question. Here is a menu of goals that a Simulationist may choose to emphasize:

\begin{itemize}
\item \textbf{Predictive validity}---Designing systems capable of generating predictions that can be validated against real-world outcomes
\item \textbf{Causal consistency}---Ensuring that system mechanics reflect theoretically sound causal relationships
\item \textbf{Empirical grounding}---Basing agent behaviors and environmental responses on empirical observations where possible
\item \textbf{Emergent complexity}---Allowing complex phenomena to emerge from the interaction of simple rules and components rather than being explicitly programmed
\end{itemize}

Within the Simulationist perspective, it's useful to distinguish between the goals of prediction and simulation itself. Prediction focuses primarily on the accuracy of forecasting future outcomes, regardless of whether the prediction gathers any insight into the underlying process. In contrast, simulation prioritizes the fidelity of the model's of real-world mechanisms. A successful simulation should ideally yield accurate predictions (demonstrating predictive validity), but the core Simulationist drive often lies in using the model as a laboratory: to explore counterfactuals, understand systematic behavior, and gain insight into the causal levers.

The Entity-Component architecture is key to achieving these goals. The engineer provides a library of components with varying trade-offs in fidelity and computational cost. The designer then composes a model at the appropriate level of abstraction for their question—from using a simple RationalActor component for a large-scale economic forecast to a more complex BeliefsAndReasoning component for a small-group study. This allows for the iterative refinement needed to ensure the system's behavior aligns with real-world data, prioritizing causal consistency and explanatory power.

\section{Discussion}

\subsection{Synthetic Data Generation: A Cross-Cutting Concern}
\label{section:syntheticData}

Synthetic data generation represents a distinct motivation for users of multi-actor generative AI systems. However, rather than constituting a fourth primary category alongside Evaluationist, Dramatist, and Simulationist motivations, we argue that data generation functions more effectively as a cross-cutting concern that permeates and influences each primary category.

Synthetic data generation is inherently purpose-driven. It serves goals such as enabling models to bootstrap their reasoning capabilities in verifiable domains \citep{zelikman2022star,guo2025deepseek,yu2023metamath}, or curating vast amounts of high-quality, targeted data to train specialized models \citep{eldan2023tinystories,abdin2024phi,abdin2024phi4}. This instrumental quality means that the design of data-generating systems will inevitably be colored by one of the three primary motivations:

\begin{itemize}
\item \textbf{Evaluationist-driven data generation} focuses on creating balanced, representative datasets suitable for benchmarking and comparative assessment of AI systems, for example by procedurally generating evaluations from templates \citep{gandhi2023understanding}.
\item \textbf{Dramatist-driven data generation} prioritizes creating rich, narratively complex interaction datasets that capture the nuances of character development and plot progression. This motivation also encompasses synthetic data designed to increase the creativity of the model's outputs \citep{xu2025direct,ismayilzada2025creative}.
\item \textbf{Simulationist-driven data generation} emphasizes producing data that accurately reflects real-world causal relationships and can be used to train models with strong predictive validity. 
\end{itemize}

\subsection{Design Trade-offs and System Purpose}

\cite{edwards2004system} argues that ``a good system is one which knows its outlook and doesn't waste any mechanics on the other two outlooks''. This principle is critical. A specific scenario cannot simultaneously serve Evaluationist, Dramatist, and Simulationist goals equally well; optimizing for one motivation inevitably involves trade-offs with the others. For instance, the flexible resolution mechanisms that serve a Dramatist by prioritizing narrative satisfaction may undermine the physical consistency prized by a Simulationist.

However, this does not mean the underlying platform must be limited to just one of the motivating outlooks. Concordia is best understood by analogy to a general-purpose engine, not a specific game. A game engine like Unity can be used to create a first-person shooter, a role-playing game, or a puzzle game. While the engine is flexible enough for any of these, any single product that tried to be all three at once would likely fail. Similarly, Concordia provides a library of tools to build a focused experience. It supports all three motivations, but for any given scenario, the designer must choose a primary goal if they are to be effective.

These trade-offs are shaped by the specific outcomes each configuration is designed to support. An Evaluationist system emphasises comparability and control, a Dramatist one centers narrative coherence and emotional significance, a Simulationist configuration prioritises causal fidelity and emergent dynamics. Each configuration reflects a choice about what kind of insight or output is most valuable. \cite{salen2003rules} describe game ``systems'' as structures that shape meaning through their design choices. In Concordia, this idea is actualized for each particular scenario in the way specific configurations of sets of components influence not just what an actor will do but also what kinds of questions the scenario is set up to explore. In this way, scenario design becomes more than mechanics---it becomes a way of enacting a particular approach to answering (for a given context) how knowledge is acquired, validated, and what constitutes valuable understanding. See also \cite{kommers2025meaning} on cultural ``thick description'' with LLMs.

The design choices made for an Evaluationist system reflect a commitment to obtaining knowledge through controlled comparison and quantifiable performance metrics. A Dramatist design prioritizes insights gained through the exploration of narrative possibilities and the emotional impact of the generated story. A Simulationist configuration, conversely, embodies a stance that seeks understanding through modeling complex interactions and observing emergent phenomena. Therefore, the chosen configuration shapes not just the system's output, but also the way in which researchers and users engage with the system to learn and gain knowledge, highlighting the deliberate choice about what kind of truth or insight the system is intended to reveal. Concordia's contribution is not to resolve these tensions but to make them configurable, allowing researchers to articulate and operationalize their modeling priorities through system design.

\subsection{Implications for Multi-Actor System Definition Language Design}

Concordia's design philosophy is to empower the user in their role as a ``game designer''. It includes templates---pre-configured prefabs---that work as appropriate starting points for users who have each of the three motivations, much like a game engine might offer a ``First-Person Shooter'' or ``Role-Playing Game'' template.

This separation of roles is directly inspired by common practice in teams using engines like Unity \citep{unity3d}. The designer workflow in Concordia mirrors that of a level designer in Unity, who can create new game entities by adding components and tweaking their parameters, often without writing new low-level code. For instance, a designer can ask: ``What if this actor had a more limited memory but a stronger sense of social obligation?'' They can answer this question by composing an actor with a specific Memory component and a SocialNorms component, then simply tuning the parameters on each.

Meanwhile, the engineer workflow is what makes this powerful designer experience possible. An engineer would have first implemented that SocialNorms component as a self-contained, reusable Python class within the Concordia framework, deliberately exposing key parameters for the designer to adjust. This decoupling ensures that designers can rapidly prototype and experiment with complex behaviors on a stable foundation, while engineers can focus on building and validating robust, foundational tools without getting entangled in the specifics of any single scenario.

\section{Conclusion}

The diverse applications of multi-actor generative AI—from scientific modeling to narrative generation and AI evaluation—demand a flexible and robust architectural foundation. The Entity-Component pattern, borrowed from game development, provides this foundation by separating the concerns of low-level engineering from high-level design. This allows a designer to rapidly prototype and configure complex scenarios by composing actors and environments from a library of reusable components. As demonstrated by the Concordia library, this approach empowers users to tailor systems to their specific Evaluationist, Dramatist, or Simulationist goals, advancing the capability and accessibility of generative multi-actor modeling.

The present iteration, v2, provides the technical means to realize the vision of generative agent-based modeling we have previously described \citep{vezhnevets2023generative, leibo2024theory}. By enabling actors who operate on complex verbal internal states through pattern-completion rather than simple reward maximization, this framework moves beyond the constraints of classical game theory and traditional agent-based models. It opens a new frontier where the lines between social science, interactive storytelling, and AI science begin to blur. By providing a common architectural foundation, Concordia offers a versatile space where Simulationists can build testable worlds, Dramatists can direct emergent narratives, and Evaluationists can forge rigorous benchmarks within the same unified framework.

Concordia is available on github at \url{https://github.com/google-deepmind/concordia}.

\bibliography{main}

\begin{thebibliography}{52}
\providecommand{\natexlab}[1]{#1}
\providecommand{\url}[1]{\texttt{#1}}
\expandafter\ifx\csname urlstyle\endcsname\relax
  \providecommand{\doi}[1]{doi: #1}\else
  \providecommand{\doi}{doi: \begingroup \urlstyle{rm}\Url}\fi

\bibitem[Abdin et~al.(2024{\natexlab{a}})Abdin, Aneja, Awadalla, Awadallah,
  Awan, Bach, Bahree, Bakhtiari, Bao, Behl, et~al.]{abdin2024phi}
M.~Abdin, J.~Aneja, H.~Awadalla, A.~Awadallah, A.~A. Awan, N.~Bach, A.~Bahree,
  A.~Bakhtiari, J.~Bao, H.~Behl, et~al.
\newblock Phi-3 technical report: A highly capable language model locally on
  your phone.
\newblock \emph{arXiv preprint arXiv:2404.14219}, 2024{\natexlab{a}}.

\bibitem[Abdin et~al.(2024{\natexlab{b}})Abdin, Aneja, Behl, Bubeck, Eldan,
  Gunasekar, Harrison, Hewett, Javaheripi, Kauffmann, et~al.]{abdin2024phi4}
M.~Abdin, J.~Aneja, H.~Behl, S.~Bubeck, R.~Eldan, S.~Gunasekar, M.~Harrison,
  R.~J. Hewett, M.~Javaheripi, P.~Kauffmann, et~al.
\newblock Phi-4 technical report.
\newblock \emph{arXiv preprint arXiv:2412.08905}, 2024{\natexlab{b}}.

\bibitem[Anthis et~al.(2025)Anthis, Liu, Richardson, Kozlowski, Koch, Evans,
  Brynjolfsson, and Bernstein]{anthis2025llm}
J.~R. Anthis, R.~Liu, S.~M. Richardson, A.~C. Kozlowski, B.~Koch, J.~Evans,
  E.~Brynjolfsson, and M.~Bernstein.
\newblock {LLM} social simulations are a promising research method.
\newblock \emph{arXiv preprint arXiv:2504.02234}, 2025.

\bibitem[Du{\'e}{\~n}ez-Guzm{\'a}n et~al.(2023)Du{\'e}{\~n}ez-Guzm{\'a}n,
  Sadedin, Wang, McKee, and Leibo]{duenez2023social}
E.~A. Du{\'e}{\~n}ez-Guzm{\'a}n, S.~Sadedin, J.~X. Wang, K.~R. McKee, and J.~Z.
  Leibo.
\newblock A social path to human-like artificial intelligence.
\newblock \emph{Nature machine intelligence}, 5\penalty0 (11):\penalty0
  1181--1188, 2023.

\bibitem[Edwards(2004)]{edwards2004system}
R.~Edwards.
\newblock System does matter.
\newblock \emph{The forge}, 28, 2004.

\bibitem[Eldan and Li(2023)]{eldan2023tinystories}
R.~Eldan and Y.~Li.
\newblock Tinystories: How small can language models be and still speak
  coherent english?
\newblock \emph{arXiv preprint arXiv:2305.07759}, 2023.

\bibitem[Fullerton(2008)]{fullerton2008game}
T.~Fullerton.
\newblock \emph{Game Design Workshop: A Playcentric Approach to Creating
  Innovative Games}.
\newblock Elsevier, 2008.

\bibitem[Gallotta et~al.(2024)Gallotta, Todd, Zammit, Earle, Liapis, Togelius,
  and Yannakakis]{gallotta2024large}
R.~Gallotta, G.~Todd, M.~Zammit, S.~Earle, A.~Liapis, J.~Togelius, and G.~N.
  Yannakakis.
\newblock Large language models and games: A survey and roadmap.
\newblock \emph{IEEE Transactions on Games}, 2024.

\bibitem[Gandhi et~al.(2023)Gandhi, Fr{\"a}nken, Gerstenberg, and
  Goodman]{gandhi2023understanding}
K.~Gandhi, J.-P. Fr{\"a}nken, T.~Gerstenberg, and N.~Goodman.
\newblock Understanding social reasoning in language models with language
  models.
\newblock \emph{Advances in Neural Information Processing Systems},
  36:\penalty0 13518--13529, 2023.

\bibitem[Gong et~al.(2023)Gong, Huang, Ma, Vo, Durante, Noda, Zheng, Zhu,
  Terzopoulos, Fei-Fei, et~al.]{gong2023mindagent}
R.~Gong, Q.~Huang, X.~Ma, H.~Vo, Z.~Durante, Y.~Noda, Z.~Zheng, S.-C. Zhu,
  D.~Terzopoulos, L.~Fei-Fei, et~al.
\newblock Mindagent: Emergent gaming interaction.
\newblock \emph{arXiv preprint arXiv:2309.09971}, 2023.

\bibitem[Gregory(2018)]{gregory2018game}
J.~Gregory.
\newblock \emph{Game engine architecture}.
\newblock AK Peters/CRC Press, 2018.

\bibitem[Grossmann et~al.(2023)Grossmann, Feinberg, Parker, Christakis,
  Tetlock, and Cunningham]{grossmann2023ai}
I.~Grossmann, M.~Feinberg, D.~C. Parker, N.~A. Christakis, P.~E. Tetlock, and
  W.~A. Cunningham.
\newblock Ai and the transformation of social science research.
\newblock \emph{Science}, 380\penalty0 (6650):\penalty0 1108--1109, 2023.

\bibitem[Guertler et~al.(2025)Guertler, Cheng, Yu, Liu, Choshen, and
  Tan]{guertler2025textarena}
L.~Guertler, B.~Cheng, S.~Yu, B.~Liu, L.~Choshen, and C.~Tan.
\newblock Textarena.
\newblock \emph{arXiv preprint arXiv:2504.11442}, 2025.

\bibitem[Guo et~al.(2025)Guo, Yang, Zhang, Song, Zhang, Xu, Zhu, Ma, Wang, Bi,
  et~al.]{guo2025deepseek}
D.~Guo, D.~Yang, H.~Zhang, J.~Song, R.~Zhang, R.~Xu, Q.~Zhu, S.~Ma, P.~Wang,
  X.~Bi, et~al.
\newblock Deepseek-r1: Incentivizing reasoning capability in llms via
  reinforcement learning.
\newblock \emph{arXiv preprint arXiv:2501.12948}, 2025.

\bibitem[Guo et~al.(2024)Guo, Chen, Wang, Chang, Pei, Chawla, Wiest, and
  Zhang]{guo2024large}
T.~Guo, X.~Chen, Y.~Wang, R.~Chang, S.~Pei, N.~V. Chawla, O.~Wiest, and
  X.~Zhang.
\newblock Large language model based multi-agents: A survey of progress and
  challenges.
\newblock \emph{arXiv preprint arXiv:2402.01680}, 2024.

\bibitem[Gygax and Cook(1989)]{gyraxcook1989dnd}
G.~Gygax and D.~Cook.
\newblock \emph{The Dungeon Master Guide, No. 2100, 2nd Edition (Advanced
  Dungeons and Dragons)}.
\newblock TSR, Inc, 1989.
\newblock ISBN 0880387297.

\bibitem[Han et~al.(2024)Han, Zhang, Yao, Jin, and Xu]{han2024llm}
S.~Han, Q.~Zhang, Y.~Yao, W.~Jin, and Z.~Xu.
\newblock {LLM} multi-agent systems: Challenges and open problems.
\newblock \emph{arXiv preprint arXiv:2402.03578}, 2024.

\bibitem[Hertz et~al.(2025)Hertz, K{\"o}ster, Janssen, and
  Leibo]{hertz2025beyond}
U.~Hertz, R.~K{\"o}ster, M.~A. Janssen, and J.~Z. Leibo.
\newblock Beyond the matrix: Experimental approaches to studying cognitive
  agents in social-ecological systems.
\newblock \emph{Cognition}, 254:\penalty0 105993, 2025.

\bibitem[Hewitt et~al.(2024)Hewitt, Ashokkumar, Ghezae, and
  Willer]{hewitt2024predicting}
L.~Hewitt, A.~Ashokkumar, I.~Ghezae, and R.~Willer.
\newblock Predicting results of social science experiments using large language
  models.
\newblock \emph{Preprint}, 2024.

\bibitem[Horton(2023)]{horton2023large}
J.~J. Horton.
\newblock Large language models as simulated economic agents: What can we learn
  from homo silicus?
\newblock Technical report, National Bureau of Economic Research, 2023.

\bibitem[Hunicke et~al.(2004)Hunicke, LeBlanc, and Zubek]{hunicke2004mda}
R.~Hunicke, M.~LeBlanc, and R.~Zubek.
\newblock Mda: A formal approach to game design and game research.
\newblock In \emph{Proceedings of the AAAI Workshop on Challenges in Game AI},
  2004.

\bibitem[Ismayilzada et~al.(2025)Ismayilzada, Laverghetta~Jr, Luchini, Patel,
  Bosselut, van~der Plas, and Beaty]{ismayilzada2025creative}
M.~Ismayilzada, A.~Laverghetta~Jr, S.~A. Luchini, R.~Patel, A.~Bosselut,
  L.~van~der Plas, and R.~Beaty.
\newblock Creative preference optimization.
\newblock \emph{arXiv preprint arXiv:2505.14442}, 2025.

\bibitem[Jenkins(2004)]{jenkins2004game}
H.~Jenkins.
\newblock Game design as narrative architecture.
\newblock In \emph{First person: New media as story, performance, and game},
  pages 118--130. MIT Press, 2004.

\bibitem[Kommers et~al.(2025)Kommers, Hemment, Antoniak, Leibo, Long, Robinson,
  and Sobey]{kommers2025meaning}
C.~Kommers, D.~Hemment, M.~Antoniak, J.~Z. Leibo, H.~Long, E.~Robinson, and
  A.~Sobey.
\newblock Meaning is not a metric: Using llms to make cultural context legible
  at scale.
\newblock \emph{arXiv preprint arXiv:2505.23785}, 2025.

\bibitem[Korsah(2025)]{korsah2025impact}
A.~Korsah.
\newblock The impact of generative artificial intelligence in game development:
  a scoping review.
\newblock 2025.

\bibitem[Kozlowski et~al.(2024)Kozlowski, Kwon, and Evans]{kozlowski2024silico}
A.~C. Kozlowski, H.~Kwon, and J.~A. Evans.
\newblock In silico sociology: forecasting covid-19 polarization with large
  language models.
\newblock \emph{arXiv preprint arXiv:2407.11190}, 2024.

\bibitem[Leibo et~al.(2024)Leibo, Vezhnevets, Diaz, Agapiou, Cunningham,
  Sunehag, Haas, Koster, Du{\'e}{\~n}ez-Guzm{\'a}n, Isaac, Piliouras, Bileschi,
  Rahwan, and Osindero]{leibo2024theory}
J.~Z. Leibo, A.~S. Vezhnevets, M.~Diaz, J.~P. Agapiou, W.~A. Cunningham,
  P.~Sunehag, J.~Haas, R.~Koster, E.~A. Du{\'e}{\~n}ez-Guzm{\'a}n, W.~S. Isaac,
  G.~Piliouras, S.~M. Bileschi, I.~Rahwan, and S.~Osindero.
\newblock A theory of appropriateness with applications to generative
  artificial intelligence.
\newblock \emph{arXiv preprint arXiv:2412.19010}, 2024.

\bibitem[Li et~al.(2023)Li, Hammoud, Itani, Khizbullin, and
  Ghanem]{li2023camel}
G.~Li, H.~Hammoud, H.~Itani, D.~Khizbullin, and B.~Ghanem.
\newblock {CAMEL}: Communicative agents for" mind" exploration of large
  language model society.
\newblock \emph{Advances in Neural Information Processing Systems},
  36:\penalty0 51991--52008, 2023.

\bibitem[Liu et~al.(2025)Liu, Guertler, Yu, Liu, Qi, Balcells, Liu, Tan, Shi,
  Lin, et~al.]{liu2025spiral}
B.~Liu, L.~Guertler, S.~Yu, Z.~Liu, P.~Qi, D.~Balcells, M.~Liu, C.~Tan, W.~Shi,
  M.~Lin, et~al.
\newblock Spiral: Self-play on zero-sum games incentivizes reasoning via
  multi-agent multi-turn reinforcement learning.
\newblock \emph{arXiv preprint arXiv:2506.24119}, 2025.

\bibitem[Mirowski et~al.(2023)Mirowski, Mathewson, Pittman, and
  Evans]{mirowski2023co}
P.~Mirowski, K.~W. Mathewson, J.~Pittman, and R.~Evans.
\newblock Co-writing screenplays and theatre scripts with language models:
  Evaluation by industry professionals.
\newblock In \emph{Proceedings of the 2023 CHI conference on human factors in
  computing systems}, pages 1--34, 2023.

\bibitem[Motwani et~al.(2024)Motwani, Smith, Das, Rafailov, Laptev, Torr,
  Pizzati, Clark, and de~Witt]{motwani2024malt}
S.~R. Motwani, C.~Smith, R.~J. Das, R.~Rafailov, I.~Laptev, P.~H. Torr,
  F.~Pizzati, R.~Clark, and C.~S. de~Witt.
\newblock Malt: Improving reasoning with multi-agent llm training.
\newblock \emph{arXiv preprint arXiv:2412.01928}, 2024.

\bibitem[Nystrom(2014)]{nystrom2014game}
R.~Nystrom.
\newblock \emph{Game programming patterns}.
\newblock Genever Benning, 2014.

\bibitem[Park et~al.(2023)Park, O'Brien, Cai, Morris, Liang, and
  Bernstein]{park2023generative}
J.~S. Park, J.~O'Brien, C.~J. Cai, M.~R. Morris, P.~Liang, and M.~S. Bernstein.
\newblock Generative agents: Interactive simulacra of human behavior.
\newblock In \emph{Proceedings of the 36th annual acm symposium on user
  interface software and technology}, pages 1--22, 2023.

\bibitem[Park et~al.(2024)Park, Zou, Shaw, Hill, Cai, Morris, Willer, Liang,
  and Bernstein]{park2024generative}
J.~S. Park, C.~Q. Zou, A.~Shaw, B.~M. Hill, C.~Cai, M.~R. Morris, R.~Willer,
  P.~Liang, and M.~S. Bernstein.
\newblock Generative agent simulations of 1,000 people.
\newblock \emph{arXiv preprint arXiv:2411.10109}, 2024.

\bibitem[Poteete et~al.(2010)Poteete, Janssen, and Ostrom]{poteete2010working}
A.~R. Poteete, M.~A. Janssen, and E.~Ostrom.
\newblock \emph{Working together: collective action, the commons, and multiple
  methods in practice}.
\newblock Princeton University Press, 2010.

\bibitem[Raffaillac and Huot(2019)]{raffaillac2019polyphony}
T.~Raffaillac and S.~Huot.
\newblock Polyphony: Programming interfaces and interactions with the
  entity-component-system model.
\newblock \emph{Proceedings of the ACM on Human-Computer Interaction},
  3\penalty0 (EICS):\penalty0 1--22, 2019.

\bibitem[Salen and Zimmerman(2003)]{salen2003rules}
K.~Salen and E.~Zimmerman.
\newblock \emph{Rules of Play: Game Design Fundamentals}.
\newblock MIT Press, Cambridge, MA, 2003.

\bibitem[Schill et~al.(2019)Schill, Anderies, Lindahl, Folke, Polasky,
  C{\'a}rdenas, Cr{\'e}pin, Janssen, Norberg, and Schl{\"u}ter]{schill2019more}
C.~Schill, J.~M. Anderies, T.~Lindahl, C.~Folke, S.~Polasky, J.~C.
  C{\'a}rdenas, A.-S. Cr{\'e}pin, M.~A. Janssen, J.~Norberg, and
  M.~Schl{\"u}ter.
\newblock A more dynamic understanding of human behaviour for the anthropocene.
\newblock \emph{Nature Sustainability}, 2\penalty0 (12):\penalty0 1075--1082,
  2019.

\bibitem[Shao et~al.(2024)Shao, Samuel, Jiang, Yang, and
  Yang]{shao2024collaborative}
Y.~Shao, V.~Samuel, Y.~Jiang, J.~Yang, and D.~Yang.
\newblock Collaborative gym: A framework for enabling and evaluating
  human-agent collaboration.
\newblock \emph{arXiv preprint arXiv:2412.15701}, 2024.

\bibitem[Silver and Sutton(2025)]{silver2025welcome}
D.~Silver and R.~S. Sutton.
\newblock Welcome to the era of experience.
\newblock \emph{Google AI}, 1, 2025.

\bibitem[Smith et~al.(2025)Smith, Abdulhai, Diaz, Tesic, Trivedi, Vezhnevets,
  Hammond, Clifton, Chang, Duéñez-Guzmán, Agapiou, Matyas, Karmon, and
  Leibo]{smith2025evaluating}
C.~Smith, M.~Abdulhai, M.~Diaz, M.~Tesic, R.~Trivedi, A.~Vezhnevets,
  L.~Hammond, J.~Clifton, M.~Chang, E.~A. Duéñez-Guzmán, J.~P. Agapiou,
  J.~Matyas, D.~Karmon, and J.~Z. Leibo.
\newblock Evaluating generalization capabilities of {LLM}-based agents in
  mixed-motive scenarios using concordia.
\newblock \emph{arxiv}, 2025.

\bibitem[Sunehag and Leibo(2025)]{sunehag2025simulation}
P.~Sunehag and J.~Z. Leibo.
\newblock Simulation streams: A programming paradigm for controlling large
  language models and building complex systems with generative ai.
\newblock \emph{arXiv preprint arXiv:2501.18668}, 2025.

\bibitem[{Unity Technologies}()]{unity3d}
{Unity Technologies}.
\newblock Unity {3D}.
\newblock URL \url{https://unity.com/}.

\bibitem[Vezhnevets et~al.(2023)Vezhnevets, Agapiou, Aharon, Ziv, Matyas,
  Du{\'e}{\~n}ez-Guzm{\'a}n, Cunningham, Osindero, Karmon, and
  Leibo]{vezhnevets2023generative}
A.~S. Vezhnevets, J.~P. Agapiou, A.~Aharon, R.~Ziv, J.~Matyas, E.~A.
  Du{\'e}{\~n}ez-Guzm{\'a}n, W.~A. Cunningham, S.~Osindero, D.~Karmon, and
  J.~Z. Leibo.
\newblock Generative agent-based modeling with actions grounded in physical,
  social, or digital space using concordia.
\newblock \emph{arXiv preprint arXiv:2312.03664}, 2023.

\bibitem[Wu et~al.(2024)Wu, Wu, Jiang, Liu, Hong, Zhao, and Zhang]{wu2024role}
W.~Wu, H.~Wu, L.~Jiang, X.~Liu, J.~Hong, H.~Zhao, and M.~Zhang.
\newblock From role-play to drama-interaction: An {LLM} solution.
\newblock \emph{arXiv preprint arXiv:2405.14231}, 2024.

\bibitem[Xu et~al.(2025)Xu, Chakraborty, Sharma, Nunes, K{\i}c{\i}man, Lu, and
  Chandra]{xu2025direct}
Y.~Xu, T.~Chakraborty, S.~Sharma, L.~Nunes, E.~K{\i}c{\i}man, S.~Lu, and
  R.~Chandra.
\newblock Direct reasoning optimization: Llms can reward and refine their own
  reasoning for open-ended tasks.
\newblock \emph{arXiv preprint arXiv:2506.13351}, 2025.

\bibitem[Yang et~al.(2025)Yang, Chai, Song, Qi, Wen, Li, Liao, Hu, Lin, Chang,
  et~al.]{yang2025survey}
Y.~Yang, H.~Chai, Y.~Song, S.~Qi, M.~Wen, N.~Li, J.~Liao, H.~Hu, J.~Lin,
  G.~Chang, et~al.
\newblock A survey of ai agent protocols.
\newblock \emph{arXiv preprint arXiv:2504.16736}, 2025.

\bibitem[Yu et~al.(2023)Yu, Jiang, Shi, Yu, Liu, Zhang, Kwok, Li, Weller, and
  Liu]{yu2023metamath}
L.~Yu, W.~Jiang, H.~Shi, J.~Yu, Z.~Liu, Y.~Zhang, J.~T. Kwok, Z.~Li, A.~Weller,
  and W.~Liu.
\newblock Metamath: Bootstrap your own mathematical questions for large
  language models.
\newblock \emph{arXiv preprint arXiv:2309.12284}, 2023.

\bibitem[Zelikman et~al.(2022)Zelikman, Wu, Mu, and Goodman]{zelikman2022star}
E.~Zelikman, Y.~Wu, J.~Mu, and N.~Goodman.
\newblock Star: Bootstrapping reasoning with reasoning.
\newblock \emph{Advances in Neural Information Processing Systems},
  35:\penalty0 15476--15488, 2022.

\bibitem[Zeng et~al.(2024)Zeng, Wu, Zhang, Wang, and Wu]{zeng2024autodefense}
Y.~Zeng, Y.~Wu, X.~Zhang, H.~Wang, and Q.~Wu.
\newblock Autodefense: Multi-agent llm defense against jailbreak attacks.
\newblock \emph{arXiv preprint arXiv:2403.04783}, 2024.

\bibitem[Zhang et~al.(2023)Zhang, Du, Shan, Zhou, Du, Tenenbaum, Shu, and
  Gan]{zhang2023building}
H.~Zhang, W.~Du, J.~Shan, Q.~Zhou, Y.~Du, J.~B. Tenenbaum, T.~Shu, and C.~Gan.
\newblock Building cooperative embodied agents modularly with large language
  models.
\newblock \emph{arXiv preprint arXiv:2307.02485}, 2023.

\bibitem[Zhou et~al.(2023)Zhou, Zhu, Mathur, Zhang, Yu, Qi, Morency, Bisk,
  Fried, Neubig, et~al.]{zhou2023sotopia}
X.~Zhou, H.~Zhu, L.~Mathur, R.~Zhang, H.~Yu, Z.~Qi, L.-P. Morency, Y.~Bisk,
  D.~Fried, G.~Neubig, et~al.
\newblock Sotopia: Interactive evaluation for social intelligence in language
  agents.
\newblock \emph{arXiv preprint arXiv:2310.11667}, 2023.

\end{thebibliography}

\end{document}